\pdfoutput=1

\documentclass[11pt]{article}

\usepackage{EMNLP2023}

\usepackage{times}
\usepackage{latexsym}

\usepackage[T1]{fontenc}

\usepackage[utf8]{inputenc}

\usepackage{microtype}

\usepackage{inconsolata}
\usepackage{graphicx}

\usepackage{amsthm}
\usepackage{amssymb}
\usepackage{amsmath}
\usepackage{booktabs}
\usepackage{multirow} 
\usepackage{tabularx} %
\usepackage{graphicx}%
\usepackage{adjustbox}
\usepackage{colortbl}
\newcommand{\ot}{\textit{o}}
\newcommand{\at}{\textit{a}}

\newcommand{\focus}{\textsc{Focus}}
\newcommand{\wechsel}{\textsc{Wechsel}}
\newcommand{\wechselen}{\textsc{Wechsel$_\textsc{En}$}}
\newcommand{\lapt}{\textsc{Lapt}}
\newcommand{\fasttext}{fastText}
\newcommand{\sparsemax}{$\mathrm{sparsemax}$}

\usepackage{hyperref}
\usepackage{tikz}
\usepackage{pifont}
\usepackage{enumitem}
\usepackage{textcomp}
\usepackage{csquotes}
\usepackage{placeins}
\usepackage{mathtools}
\usepackage{bm}
\usepackage{esvect}
\usepackage{listings,multicol}
\DeclareMathOperator*{\argmin}{\arg\!\min}

\let\oldding\ding%
\renewcommand{\ding}[2][1]{\scalebox{#1}{\oldding{#2}}}%

\usepackage{etoolbox}
\makeatletter
\patchcmd{\hyper@makecurrent}{%
    \ifx\Hy@param\Hy@chapterstring
        \let\Hy@param\Hy@chapapp
    \fi
}{%
    \iftoggle{inappendix}{%
        \@checkappendixparam{chapter}%
        \@checkappendixparam{section}%
        \@checkappendixparam{subsection}%
        \@checkappendixparam{subsubsection}%
        \@checkappendixparam{paragraph}%
        \@checkappendixparam{subparagraph}%
    }{}%
}{}{\errmessage{failed to patch}}

\newcommand*{\@checkappendixparam}[1]{%
    \def\@checkappendixparamtmp{#1}%
    \ifx\Hy@param\@checkappendixparamtmp
        \let\Hy@param\Hy@appendixstring
    \fi
}
\makeatletter

\newtoggle{inappendix}
\togglefalse{inappendix}

\apptocmd{\appendix}{\toggletrue{inappendix}}{}{\errmessage{failed to patch}}

\newcommand{\B}{\fontseries{b}\selectfont}

\newcommand*{\In}{%
  \mathrel{%
    \mathpalette\@In{\in}%
  }%
}

\usepackage{todonotes}

\usepackage{titlesec}

\title{\textsc{Focus}: Effective Embedding Initialization for \\Monolingual Specialization of Multilingual Models}
\hypersetup{
  pdftitle = {FOCUS: Effective Embedding Initialization for Monolingual Specialization of Multilingual Models},
  pdfauthor = {Konstantin Dobler, Gerard de Melo}
}

  \author{Konstantin Dobler \and Gerard de Melo\\
  Hasso Plattner Institute / University of Potsdam \\
  \texttt{\{konstantin.dobler, gerard.demelo\}@hpi.de}}

\begin{document}
\maketitle
\begin{abstract}

Using model weights pretrained on a high-resource language as a warm start can reduce the need for data and compute to obtain high-quality language models for other, especially low-resource, languages. However, if we want to use a new tokenizer specialized for the target language, we cannot transfer the source model's embedding matrix. In this paper, we propose \mbox{\focus{} --} \underline{\textbf{F}}ast \underline{\textbf{O}}verlapping Token \underline{\textbf{C}}ombinations \underline{\textbf{U}}sing \underline{\textbf{S}}parsemax, a novel embedding initialization method that initializes the embedding matrix effectively for a new tokenizer based on information in the source model's embedding matrix.
\focus{} represents newly added tokens as combinations of tokens in the overlap of the source and target vocabularies.
The overlapping tokens are selected based on semantic similarity in an auxiliary static token embedding space. We focus our study on using the multilingual XLM-R as a source model and empirically show that \focus{} outperforms random initialization and previous work in language modeling and on a range of downstream tasks (NLI, QA, and NER).
We publish our checkpoints and code on GitHub.\footnote{\url{https://github.com/konstantinjdobler/focus}}
\end{abstract}

\protected\def\verythinspace{%
  \ifmmode
    \mskip0.5\thinmuskip
  \else
    \ifhmode
      \kern0.08334em
    \fi
  \fi
}
\section{Introduction}
\label{sec:intro}

Research on large language models is advancing rapidly with powerful new models being published at a break-neck pace~\citep[\textit{e.g.,}][]{
zeng2022glm,le-scao-etal-2022-language,touvron2023llama}.
Although multilingual models have been released, many of the world's languages are not covered. Multilingual models have also been shown to have subpar performance on under-resourced languages~\citep{wu-dredze-2020-languages}.
Therefore, it is crucial to develop methods that harness these advances and make them available for further languages, especially low-resource ones.

A promising line of work in this regard focuses on crosslingual transfer of Transformer models pretrained on
high-resource languages.
Crosslingual transfer directly copies the pretrained weights in the Transformer layers to the target language model.
Subsequently, the model is further adapted to the target language by continued pretraining on unlabeled target language text using the original self-supervised pretraining objective. This sort of training regimen is also known as language adaptive pretraining~(\lapt{}; \citealp{language-adaptive-pretraing}). 
\begin{figure}
    \centering
    \includegraphics[width=\linewidth]{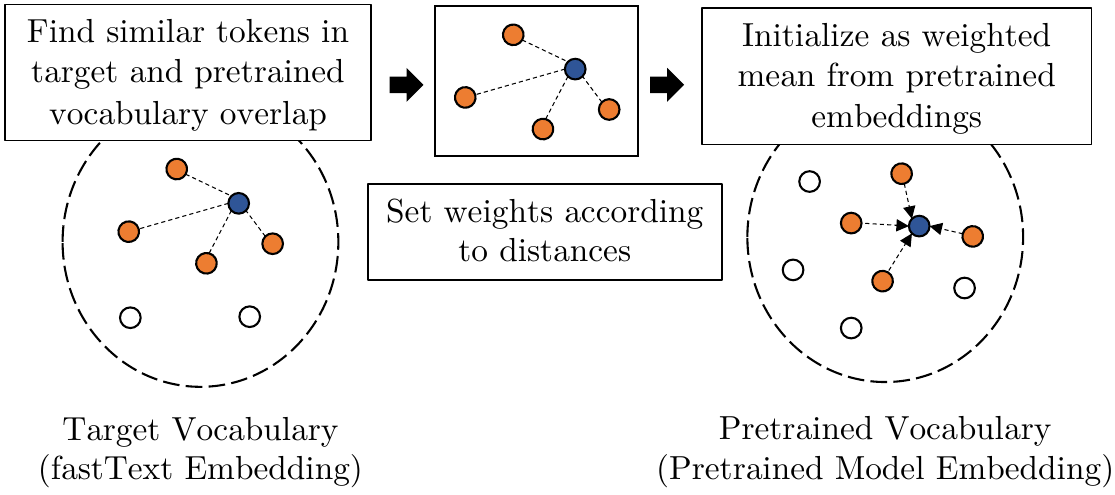}
    \vspace*{-7mm}
    \caption{Illustration of \focus{}'s initialization strategy for embeddings of new tokens (blue dot): Find similar tokens (orange dots) in an auxiliary \fasttext{} embedding space; then initialize the new token as their weighted mean in the pretrained embedding space.
    }
    \label{fig:focus-detail}
    \vspace{-4mm}
\end{figure}

However, the 
pretrained model's embedding matrix cannot be directly transferred if we use a new tokenizer for the target language~\cite{artetxe-etal-2020-cross,de-vries-nissim-2021-gpt2-recycle}. Using appropriate tokenizers has been shown to be important for the model's performance on downstream tasks~\cite{rust-etal-2021-good} and is crucial if the source and target language use different scripts.

We present \focus{}, an embedding initialization method that allows us to transfer information from the source model's pretrained embedding matrix to a new embedding matrix for the target language's tokenizer. \focus{} is illustrated in \autoref{fig:focus-detail}.
The key idea is to use overlapping tokens between both tokenizers as anchor points and represent new target language tokens as a weighted mean of overlapping tokens' embeddings. 
This enables us to initialize the new embedding matrix in the same semantic space as the pretrained embedding matrix.
We empirically show in extensive experiments across a range of different high-resource and low-resource target languages that \focus{} outperforms
various strong baselines
both in language modeling as well as on downstream tasks (Natural Language Inference, Question Answering, and Named Entity Recognition). 

In our experiments, we focus on the multilingual XLM-R~\cite{xlmr} as a source model and specialize it for a single language.
\focus{} is particularly well-positioned to take advantage of multilingual source models due to their larger vocabulary and the fact that they have already been pretrained to a certain extent on many potential target languages.
Additionally, we show that \focus{} still improves significantly over random initialization even if only minimal vocabulary overlap is available.\footnote{We study a minimal overlap consisting exclusively of tokens that are numbers, punctuation, and whitespace.}

In previous work, a common approach has been to adapt multilingual models to target languages while simply keeping or extending the original vocabulary~\cite{wang-etal-2019-improving,language-adaptive-pretraing,wang-etal-2020-extending,chau-smith-2021-specializing}. When extending the vocabulary, \focus{} can also be applied to initialize embeddings just for the new tokens.
However, we advocate considering the setting of full vocabulary replacement.
Only a fraction of the multilingual vocabulary is actually used for any single language, so by fully replacing the large multilingual vocabulary with a language-specific smaller vocabulary we can enable faster training times and smaller models.
\mbox{XLM-R's} vocabulary has 250k tokens, and replacing this with a language-specific 50k token vocabulary reduces the model size quite dramatically, by over 55\%.\footnote{Comparing 278 million parameters at $\approx$ 1.1 GB to 124 million parameters at $\approx$ 0.5 GB.}
In our experiments, training with a language-specific 50k token vocabulary is 40\% faster than extending the original 250k token vocabulary.\footnote{Comparing a 50k token vocabulary to an extended 273k token vocabulary with HuggingFace \texttt{tokenizers} and \texttt{transformers} on two Nvidia A100 80GB GPUs.}
We summarize the contributions of our paper as follows:
\begin{itemize}[leftmargin=10pt,rightmargin=1pt,itemsep=0pt,topsep=5pt]
    \item We propose \focus{}, a novel embedding initialization method that effectively transfers knowledge from a pretrained embedding matrix to one for a new, language-specific tokenizer.
    \item We empirically verify the effectiveness of \focus{} for language modeling and downstream tasks in extensive experiments using XLM-R as a source model on a range of high- and low-resource languages.
    \item We further show that \focus{} is effective also when the target language was not part of the source models' pretraining or only a minimal vocabulary overlap is available.

  \end{itemize}

\section{\focus{}}

Our goal is to initialize embeddings for tokens in a new, language-specific target vocabulary in the same semantic space as the source model's embeddings. In this study, we mainly focus on the multilingual source model XLM-R although \focus{} can in principle also be applied to monolingual source models.\footnote{In \autoref{sec:results}, we also conduct experiments in a setting with almost no vocabulary overlap, showing that \focus{} can be used to transfer a model to previously unseen languages.}
We copy all embeddings of shared tokens between source and target tokenizer for our new embedding matrix.
If the target language was already part of the source model's pretraining corpus, this takes advantage of target language tokens with pretrained embeddings in the source model's tokenizer. 
In any case, we take advantage of shared named entities, symbols, numbers, punctuation, and shared words resulting from code-switching between the target and pretrained vocabularies.
Additional target language tokens not present in the source model are represented as a %
linear combination of 
embeddings of semantically similar shared tokens.
Unlike previous work on embedding initialization, this requires neither bilingual dictionaries nor an alignment of embedding spaces across different languages~\cite{minixhofer-etal-2022-wechsel,zeng2022greenplm}. 
Next, we formally describe \focus{}.

\paragraph{Details of \focus{}.}
We obtain as input a source vocabulary $\mathbb{V}^\text{s}$ with pretrained embeddings $\mathbf{E}^\text{s}$ and a target vocabulary $\mathbb{V}^\text{t}$ with embeddings $\mathbf{E}^\text{t}$, which we seek to initialize. 
The target vocabulary  $\mathbb{V}^\text{t}$ is obtained by training a tokenizer on monolingual text in the target language.
We use
$\vv{\bm{e}}_i^s$ and $\vv{\bm{e}}_i^t$ 
to denote embeddings for individual tokens in  $\mathbf{E}^\text{s}$  and $\mathbf{E}^\text{t}$, respectively.
We denote the set of overlapping tokens as $\mathbb{O} = \mathbb{V}^\text{s} \cap \mathbb{V}^\text{t}$. 
For each overlapping token we can copy the pretrained embedding over into our target embedding matrix:
\begin{align}
    \forall \ot{} \in \raisebox{-0.8pt}{$\mathbb{O}$}:~~  \vv{\bm{e}}_o^t = \vv{\bm{e}}_o^s.
\end{align}
Note that we make an assumption here: tokens that are part of the overlap $\mathbb{O}$ have sufficiently similar semantics in our source and target vocabularies.
For multilingual source models, we can exploit already existing tokens from the target language.
Otherwise, this will obviously not always be the case\footnote{Consider words known as \emph{false friends}: words with the same spelling but different meanings across languages.}, but
through common named entities, code-switched tokens, symbols, numbers, and punctuation 
this assumption will hold reasonably often. We provide an in-depth analysis in Appendix~\ref{app:fasttext:par:vocab}.

Finding an initialization in the same semantic space as the pretrained embeddings is not as easy for the set of non-overlapping (``additional'') tokens $\mathbb{A} = \mathbb{V}^\text{t} \setminus \mathbb{O}$.
To initialize embeddings for the additional tokens, we first train auxiliary embeddings
$\mathbf{X}$ for all target tokens $\mathbb{V}^\text{t}$ (i.e., both $\mathbb{O}$ and $\mathbb{A}$).\footnote{In principle, we could instead also obtain token embeddings from already pretrained \fasttext{} word embeddings following \citet{minixhofer-etal-2022-wechsel}.
We show in \autoref{sec:results} that training \fasttext{} directly at the token level provides a better initialization.}
In our experiments, we apply \fasttext{} on unlabeled target language data pre-tokenized with the target tokenizer for $\mathbb{V}^\text{t}$.
Individual embeddings in \textbf{X} are denoted by $\vv{\bm{x}}_i$. 
Next, we compute the pairwise cosine similarities between the auxiliary embeddings $\vv{\bm{x}}_i$ of tokens in $\mathbb{A}$ and $\mathbb{O}$ so that for any $a \in \mathbb{A}$:
\newcommand{\transpose}{\intercal}
\begin{align}
\vv{\bm{c}}_a = [ \verythinspace \mathrm{sim}\verythinspace(\at{}, o_1),\verythinspace \ldots , \verythinspace\mathrm{sim}\verythinspace(\at{}, o_n) \verythinspace ]
\end{align}
where $o_i$ is the overlapping token at index $i$ and:
\begin{align}
    \mathrm{sim\verythinspace(\at{}, \ot{})} := 
    \frac{{\vv{\bm{x}}_a}\cdot \vv{\bm{x}}_o}
    {\|{\vv{\bm{x}}_a }\| \|\vv{\bm{x}}_o\|}.
\end{align}
We convert the similarity scores $\vv{\bm{c}}_a$ to weights 
by applying {\sparsemax{}}~\cite{martins2016sparsemax} over each $\vv{\bm{c}}_a$.
$\mathrm{Sparsemax}$ is a sparse variant of softmax that assigns zero probability mass to low-probability elements, which has previously been used by \citet{tran2020ramen} in a similar setting. 
Using {\sparsemax{}}
has the advantage of being able to dynamically accommodate different degrees of skew in the similarity distribution. In some cases we might have only one or two very similar tokens, in other cases, we might have significantly more.
Accordingly, the weights $\vv{\bm{w}}_a$ of the overlapping tokens are:
\newcommand{\unaryminus}{\hspace{1pt}\scalebox{0.6}[1.0]{\( - \)}\hspace{1pt}}
\newcommand{\unaryminustwo}{\hspace{0.5pt}\scalebox{0.5}[1.0]{\( - \)}\hspace{0.5pt}}
\begin{equation}
\vv{\bm{w}}_a\! = \mathrm{sparsemax}(\vv{\bm{c}}_a)=\argmin_{\vv{\bm{p}} \in \Delta}\|\vv{\bm{p}}\unaryminus \vv{\bm{c}}_a \|^2
\end{equation}
with $\Delta$ denoting the ($\lvert\mathbb{O}\rvert \unaryminustwo 1$)\unaryminus dimensional probability simplex, i.e.,
\begin{equation}
    \Delta := \{\vv{\bm{p}}  \in \mathbb{R}^{\lvert\mathbb{O}\rvert} \mid \vv{\bm{1}} \cdot \vv{\bm{p}}  = 1, \vv{\bm{p}} \geq \bm{0}\}.
\end{equation}
We then initialize the target embeddings for each additional token \at{} as a weighted mean over pretrained embeddings of the overlapping tokens from \textbf{E}$^\text{s}$, with the weights given by $\vv{\bm{w}}_a$. Due to \sparsemax{}, most of the elements in each $\vv{\bm{w}}_a$ will be zero. Note that we use the pretrained embeddings \textbf{E}$^\text{s}$ instead of the auxiliary embeddings \textbf{X}, as only the pretrained embeddings are in the same semantic space as the rest of the transferred Transformer layers. Therefore:
\begin{align}
      \forall \at{} \in \mathbb{A}:~~ \vv{\bm{e}}_a^t =  \sum_{\ot{}\in \mathbb{O}} w_{\at{},\ot{}}  \, \vv{\bm{e}}_o^s.
\end{align}
\paragraph{Summary.} \focus{} uses cheap and fast-to-train static embeddings for tokens in the target vocabulary to select semantically similar overlapping tokens for each additional target token. The {pretrained} embeddings of the overlapping tokens are then used to initialize embeddings for the additional target tokens. In Appendix~\ref{app:further-focus-details}, we provide further implementation details as well as a detailed analysis of the different types of overlapping tokens we encountered in our experiments.

\section{Experimental Setup}
\label{sec:experimental-setup}

We perform experiments using XLM-R as our multilingual source model, due to its popularity and widespread use.\footnote{{As of October 2023, XLM-R has 12.2 million downloads per month on the HuggingFace Hub.}} We use the \texttt{base} variant for all experiments. Our language-specific tokenizers are trained in the same way as XLM-R for comparability, specifically SentencePiece tokenization~\cite{sentencepiece} with the Unigram algorithm~\cite{unigram2018kudo}. We use HuggingFace \texttt{tokenizers} and a vocabulary size of 50k tokens for all languages.

\subsection{Baselines}
To evaluate \focus{}, we compare against multiple strong baselines for embedding initialization as well as other methods of adapting XLM-R to a target language. We always transfer all layers of XLM-R, except for the embedding. \citet{minixhofer-etal-2022-wechsel} already demonstrate the superiority of this over random initialization of all weights, so we do not compare against the weak baseline of training a model completely from scratch.

\paragraph{XLM-R with the original vocabulary.} We report results of using XLM-R off-the-shelf without language-adaptive pretraining (\lapt{}) as well as after adapting XLM-R to the target language with the original vocabulary kept as-is.

\paragraph{Random Initialization.}
For vocabulary replacement with a language-specific tokenizer and random embedding initialization, we copy the original pretrained embeddings following \citet{zoph-etal-2016-transfer}. 
This randomly maps pretrained embeddings to tokens in the new vocabulary and performed slightly better than other types of random initialization in preliminary experiments.\footnote{We refer to this as random initialization. Random mapping performed slightly better than initialization from a normal or uniform distribution.} In this case, we also consider the variant of training just the embeddings for an additional 20\% of training steps before unfreezing the rest of the network. \Citet{de-vries-nissim-2021-gpt2-recycle} note that this allows the new embeddings to adapt to the transferred Transformer layers to prevent catastrophic forgetting.
Therefore, this strong baseline is trained 20\% longer than other methods.

\paragraph{\wechsel{}.}
We additionally compare against using \wechsel{} \cite{minixhofer-etal-2022-wechsel} to initialize the embedding matrix for the language-specific tokenizer. 
\wechsel{} is a method for embedding initialization originally designed for transferring monolingual source models. It relies on aligning pretrained word embeddings for the source and target languages using the Orthogonal Procrustes method~\cite{orthogonalprocrustes_schonemann} with bilingual dictionaries as seed data. 
Then, each source and target token is embedded into the same semantic space using the out-of-vocabulary method of \fasttext{}, resulting in aligned static token embeddings for both languages. 

To faithfully apply \wechsel{} with a multilingual source model, we would need to provide a word embedding space for all the languages that are part of the multilingual models' pretraining corpus.
Also, gathering bilingual dictionaries from all source languages to the target language would become a challenge.
Instead, we apply \wechsel{} as-is using only pretrained English \fasttext{} word embeddings for the source model. 
This effectively assumes that all pretrained source token embeddings are English, which is a rough but not entirely unreasonable assumption given the predominance of English over other languages in the pretraining corpus of XLM-R. 
We can further commit to this assumption by deleting all non-English tokens from the pretrained vocabulary before applying \wechsel{}, which we dub \wechselen{}.
This yields an initialization method similar to the mixture mapping method proposed by \citet{wang-etal-2019-improving}.

\paragraph{Vocabulary Extension.}
We also run experiments with vocabulary extension following \citet{wang-etal-2020-extending} by extending with the top 30k tokens of the language-specific tokenizer as well as using \focus{} to initialize embeddings for the extended tokens.

\begin{table}

    \begin{tabular}{lr}
    \toprule

     Language & Dataset Size (GB) \\
    \midrule
    German & 18 GB\\
    Arabic &  5.4 GB\\
    Kiswahili & 0.3 GB\\
    Hausa & 0.06 GB \\
    isiXhosa & 0.03 GB\\
    \bottomrule
    \end{tabular}
    \centering
    \caption{Size of datasets from CC100 used for \lapt{}.}
    \label{tab:pretrain-data-sizes}
    \vspace{-3mm}
\end{table}

\subsection{Language-Adaptive Pretraining (\lapt{})}
For \lapt{}, we use the same self-supervised Masked Language Modeling (\textsc{MLM}) objective as in the original pretraining of XLM-R. We use the CC100 corpus to obtain unlabeled text in our target languages, which was also already used for the pretraining of \mbox{XLM-R}~\cite{xlmr}. 
Therefore, we do not introduce any new unseen data. We show dataset sizes for our target languages in \autoref{tab:pretrain-data-sizes}.
We use the same hyperparameters for all languages, as detailed in Appendix~\ref{app:hyperparams:par:further-pretraining}. In particular, we use a batch size of 128 with chunked sequences of 256 tokens and train our models on 50 million samples (resulting in a total of 12.8 billion training tokens and 390k optimizer steps). 

\subsection{Evaluation}
\label{sec:experiment-setup-eval}
We also evaluate our models on downstream tasks in their respective target languages.
We perform downstream task evaluation on five languages: German, Arabic, Kiswahili, isiXhosa, and Hausa. 
They were chosen to provide a mix of high-, medium- and low-resource languages, typological and script diversity while satisfying the practical constraints of available evaluation datasets. 
We refer to German as high-resource, Arabic and Kiswahili as medium-resource, and isiXhosa and Hausa as low-resource languages.

We use the translated training sets of XNLI~\cite{conneau-etal-2018-xnli} to evaluate Natural Language Inference (NLI) in the translate--train setting. To evaluate Question Answering, we use GermanQuAD~(for German, \citealp{moller-etal-2021-germanquad}) and TyDiQA GoldP (for Swahili and Arabic; \citealp{clark-etal-2020-tydi}).
We perform Named Entity Recognition (NER) experiments using the balanced \texttt{train}--\texttt{dev}--\texttt{test} split of WikiANN~\cite{rahimi-etal-2019-massively, pan-etal-2017-cross}. Additionally, we evaluate NER for German on the GermEval2014 dataset~\cite{benikova-etal-2014-nosta} and for Swahili, Hausa, and isiXhosa using MasakhaNERv2~\citep{adelani-etal-2022-masakhaner}.
If there is no dedicated \texttt{dev} split, we construct our own with a random sample of 10\% of the training data. We perform model selection on the \texttt{dev} split and report the selected checkpoint's result on the test set. We report accuracy for XNLI and F$_1$-scores otherwise. We run all experiments five times with different random seeds and report the mean and standard deviation. Hyperparameters for all evaluation tasks are given in Appendix~\ref{app:hyperparams:par:downstream}. 

Furthermore, we evaluate the initialization performance of \focus{} without further training measured by the MLM loss on a held out set on five additional very low-resource languages (Scottish Gaelic, Luxembourgish, Cebuano, Samoan, and Hmong). For these languages, we use mC4~\cite{raffelExploringLimitsTransfer2020} and OSCARv23.01~\cite{abadji-etal-2022-towards} as additional data sources for unlabeled text.

\section{Results}
\label{sec:results}

\begin{table*}
    \centering
    \resizebox{1\linewidth}{!}{
    \begin{tabular}{@{}l c c c c c c c c c ccc@{}}
    \toprule
    & \multicolumn{4}{c}{XNLI (translate-train)} && \multicolumn{4}{c}{GermanQuAD / TyDiQA}\\
    \cmidrule(lr){2-5}
    \cmidrule(lr){7-10}
    Method  & German & Arabic & Kiswahili & Avg. && German & Arabic & Kiswahili & Avg.\\
    \midrule
XLM-R (original vocab)\\
- off-the-shelf & 78.8 \small$\pm$ 0.3 & 74.8 \small$\pm$ 0.4 & 69.1 \small$\pm$ 0.4 & 74.2 &&  \underline{\B71.3} \small$\pm$ 0.4 & 78.4 \small$\pm$ 0.9 & 73.9 \small$\pm$ 1.4 & 74.5\\
- \lapt{}  & \underline{{\B78.9}} \small$\pm$ 0.4 & {\B75.1} \small$\pm$ 0.6 & {\B72.4} \small$\pm$ 0.4 & \B75.5 &&  70.5 \small$\pm$ 0.8 & {\B78.9} \small$\pm$ 0.5 & {\B75.8} \small$\pm$ 1.0 & \B75.0\\
\cmidrule{1-10}
XLM-R (replaced vocab)\\
- Random + \lapt{}\textsuperscript{\textdagger} & 77.6 \small$\pm$ 0.4 & 74.6 \small$\pm$ 0.4 & 71.2 \small$\pm$ 0.3 & 74.5 &&  69.1 \small$\pm$ 0.7 & 79.3 \small$\pm$ 0.6 & 74.2 \small$\pm$ 1.0 & 74.2\\
- \wechselen{} + \lapt{} & 77.7 \small$\pm$ 0.5 & 75.4 \small$\pm$ 0.3 & 72.0 \small$\pm$ 0.2 & 75.0 &&  71.0 \small$\pm$ 0.4 & 79.3 \small$\pm$ 1.0 & 75.2 \small$\pm$ 0.7 & 75.2\\
- \wechsel{} + \lapt{} & 78.2 \small$\pm$ 0.2 & 76.0 \small$\pm$ 0.2 & 72.3 \small$\pm$ 0.3 & 75.5 &&  70.5 \small$\pm$ 0.5 & \underline{{\B79.4}} \small$\pm$ 0.9 & 75.5 \small$\pm$ 1.5 & 75.1\\
- \focus{} + \lapt{} & {\B78.3} \small$\pm$ 0.6 & \underline{{\B76.5}} \small$\pm$ 0.4 & \underline{{\B72.9}} \small$\pm$ 0.5 &  \underline{\B75.9} &&  \underline{{\B71.3}} \small$\pm$ 0.2 & 79.1 \small$\pm$ 0.4 & {\B76.5} \small$\pm$ 1.5 & \underline{\B75.6}\\
\cmidrule{1-10}
XLM-R (extended vocab)\\
- Random + \lapt{} & 77.7 \small$\pm$ 0.6 & 75.2 \small$\pm$ 0.6 & 71.8 \small$\pm$ 0.4 & 74.9 &&  {\B69.8} \small$\pm$ 0.6 & 77.7 \small$\pm$ 0.6 & 76.3 \small$\pm$ 0.8 & 74.6\\
- \focus{} + \lapt{} & {\B78.0} \small$\pm$ 0.4 & {\B75.5} \small$\pm$ 0.4 & {\B72.1} \small$\pm$ 0.2 & \B75.2 &&  69.5 \small$\pm$ 0.3 & {\B77.8} \small$\pm$ 1.0 & \underline{{\B77.0}} \small$\pm$ 0.6 & \B74.7\\
    \bottomrule
    \end{tabular}
    }
    \caption{Results on Natural Language Inference and Question Answering tasks. Details on the datasets used for evaluation are given in \autoref{sec:experiment-setup-eval}. We \textbf{bold} the best result in each section and \underline{underline} the overall best result. \lapt{} is short for language-adaptive pretraining; we perform \lapt{} for 50 million samples on unlabeled target texts. \textsuperscript{\textdagger}: For random initialization, we train just the embeddings for an additional 20\% of training steps before full \lapt{} to create a stronger baseline.}
    \label{tab:xnli-qa}
\end{table*}

\begin{table*}
    \centering
    \resizebox{1\linewidth}{!}{
    \begin{tabular}{@{}l c c c c c c c c c c @{}}
    \toprule
    & \multicolumn{4}{c}{WikiANN} && \multicolumn{5}{c}{GermEval14 / MasakhaNERv2}\\
    \cmidrule(lr){2-5}
    \cmidrule(lr){7-11}
    Method  & German & Arabic & Kiswahili & Avg. && German & Kiswahili & Hausa & isiXhosa & Avg.\\
    \midrule
- off-the-shelf & 86.3 \small$\pm$ 0.2 & 85.7 \small$\pm$ 0.3 & 86.6 \small$\pm$ 0.5 & 86.2 &&  85.6 \small$\pm$ 0.3 & 92.0 \small$\pm$ 0.1 & 84.2 \small$\pm$ 0.5 & 85.5 \small$\pm$ 0.3 & 87.1\\
- \lapt{}  & \underline{{\B86.7}} \small$\pm$ 0.1 & {\B87.1} \small$\pm$ 0.1 & {\B86.9} \small$\pm$ 0.6 & \B86.9 &&  \underline{{\B86.8}} \small$\pm$ 0.2 & {\B92.5} \small$\pm$ 0.2 & {\B85.6} \small$\pm$ 0.4 & {\B88.3} \small$\pm$ 0.2 & \B88.3\\
\cmidrule{1-11}
XLM-R (replaced vocab)\\
- Random + \lapt{}\textsuperscript{\textdagger}\ & 86.0 \small$\pm$ 0.1 & 87.5 \small$\pm$ 0.1 & 85.8 \small$\pm$ 0.5 & 86.4 &&  85.9 \small$\pm$ 0.3 & 92.3 \small$\pm$ 0.2 & 85.0 \small$\pm$ 0.3 & 87.4 \small$\pm$ 0.2 & 87.8\\
- \wechselen{} + \lapt{} & 86.4 \small$\pm$ 0.1 & 87.8 \small$\pm$ 0.1 & 86.6 \small$\pm$ 0.9 & 87.0 &&  86.4 \small$\pm$ 0.2 & 92.3 \small$\pm$ 0.1 & --\textsuperscript{\textdaggerdbl} & --\textsuperscript{\textdaggerdbl} & --\\
- \wechsel{} + \lapt{} & 86.5 \small$\pm$ 0.2 & \underline{{\B87.9}} \small$\pm$ 0.3 & \underline{{\B87.4}} \small$\pm$ 0.6 & \underline{\B87.3} &&  {\B86.7} \small$\pm$ 0.1 & 92.2 \small$\pm$ 0.1 & --\textsuperscript{\textdaggerdbl} & --\textsuperscript{\textdaggerdbl} & --\\
- \focus{} + \lapt{} & {\B86.6} \small$\pm$ 0.2 & \underline{\B87.9} \small$\pm$ 0.1 & 86.9 \small$\pm$ 0.4 & 87.1 &&  86.6 \small$\pm$ 0.0 & \underline{{\B92.6}} \small$\pm$ 0.1 & \underline{{\B86.0}} \small$\pm$ 0.4 & \underline{{\B88.5}} \small$\pm$ 0.4 & \underline{\B88.4}\\
\cmidrule{1-11}
XLM-R (extended vocab)\\
- Random + \lapt{} & 85.6 \small$\pm$ 0.2 & 85.2 \small$\pm$ 0.3 & {\B86.2} \small$\pm$ 0.7 & 85.6 &&  85.4 \small$\pm$ 0.3 & 92.0 \small$\pm$ 0.2 & 84.1 \small$\pm$ 0.2 & 87.2 \small$\pm$ 0.4 & 87.5\\
- \focus{} + \lapt{} & {\B86.0} \small$\pm$ 0.1 & {\B85.3} \small$\pm$ 0.3 & {\B86.2} \small$\pm$ 0.3 & \B85.8 &&  {\B85.6} \small$\pm$ 0.2 & {\B92.1} \small$\pm$ 0.2 & {\B84.9} \small$\pm$ 0.4 & {\B87.7} \small$\pm$ 0.3 & \B87.9\\

    \bottomrule
    \end{tabular}
    }
    \caption{Results on Named Entity Recognition (NER) tasks. Details on the datasets used for evaluation are given in \autoref{sec:experiment-setup-eval}. We \textbf{bold} the best result in each section and \underline{underline} the overall best result.
     \textsuperscript{\textdagger}: For random initialization, we train just the embeddings for an additional 20\% of training steps before full \lapt{} to create a stronger baseline.
    --\textsuperscript{\textdaggerdbl}: Languages not covered by the pretrained \fasttext{} word embeddings used by \wechsel{}.
    }
    \label{tab:ner}
\end{table*}

\begin{figure*}
\centering

     \hfill
  
         \includegraphics[width=.33\linewidth]{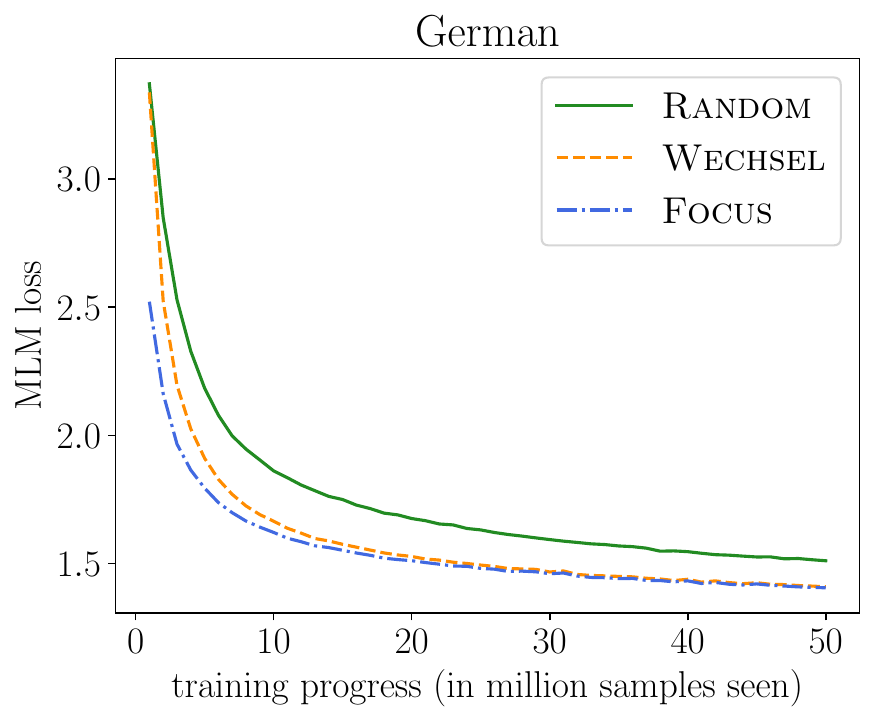}
     \hfill
         \includegraphics[width=.32\linewidth]{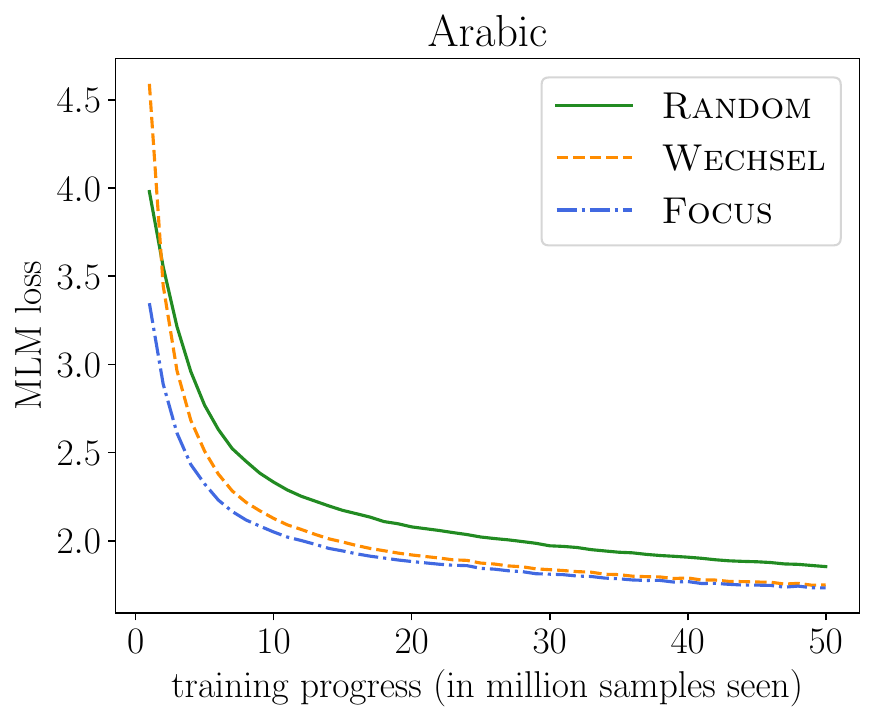}
     \hfill
         \includegraphics[width=.32\linewidth]{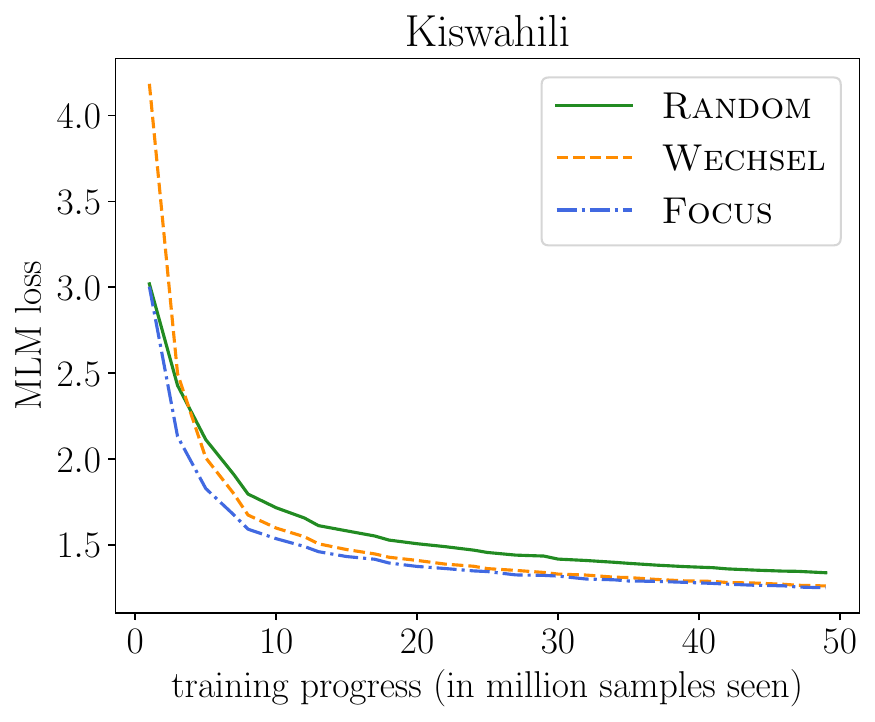}
     \hfill

        \caption{Masked Language Modeling (MLM) loss of different methods for vocabulary replacement over the course of further pretraining (\lapt{}), evaluated on a held out set. The first data point is logged at 1 million samples. For random initialization, we plot only the second stage, i.e., after already training just the embeddings for 10 million samples. This allows us to compare \focus{} and \wechsel{} embedding initialization directly with gradient descent training of the embeddings. } 
      
\label{fig:mlm-loss-comparison}
\end{figure*}

 \begin{table*}[t]
    \centering
    \resizebox{1\linewidth}{!}{
    \begin{tabular}{@{}lccccccccccc@{}}
    \toprule

      & \multicolumn{6}{c}{Part of XLM-R's pretraining} &&
      \multicolumn{4}{c}{Not Part of XLM-R's pretraining} \\
    \cmidrule(lr){2-7}
    \cmidrule(lr){9-12}
     Data source $\rightarrow$&  \multicolumn{6}{c}{CC100}&&    \multicolumn{2}{c}{OSCARv23.01} &  \multicolumn{2}{c}{mC4} \\
    \cmidrule(r){1-1}
    \cmidrule(lr){2-7}
    \cmidrule(lr){9-10} 
    \cmidrule(lr){11-12}

    Initialization $\downarrow$& German & Arabic & Kiswahili & isiXhosa & Hausa & Scott. Gaelic && Luxembourgish & Cebuano & Samoan & Hmong\\
    \midrule
Random & 24.0 & 24.1 & 24.2 & 25.5 & 23.7 & 22.8 && 24.4 & 22.5 & 21.9 & 22.8\\
\cmidrule{1-12}
\focus{} & \B4.0 & \B5.2 & \B4.8 & \B7.6 & \B6.0 & \B 6.1 && \B 8.2 & \B 6.3 & \B4.9 & \B 5.7\\
- Word-\fasttext{} & 4.3 & 5.5 & 6.0 & --\textsuperscript{\textdagger} & --\textsuperscript{\textdagger} & 7.5 && 8.7 & 6.7 & --\textsuperscript{\textdagger} & --\textsuperscript{\textdagger}  \\
- \text{Symbolic Overlap} & 10.6 & 10.6 & 10.7 & 10.4 & 10.4 & 10.4 && 10.2 & 9.7 & 8.4 & 8.1 \\
\cmidrule{1-12}
\wechsel{} & 8.3 & 9.8 & 11.2 & --\textsuperscript{\textdagger} & --\textsuperscript{\textdagger}  & 11.1 && 10.9 & 10.1 & --\textsuperscript{\textdagger} & --\textsuperscript{\textdagger}\\
    \bottomrule
    \end{tabular}
    }
    \caption{MLM loss on a held-out set immediately after initialization (no training performed) with full vocabulary replacement. We use the same vocabulary for all methods in a single language. Symbolic Overlap restricts overlapping tokens to numbers, punctuation, or whitespace. Word-\fasttext{} uses \wechsel{}'s method of turning pretrained word embeddings into token embeddings instead of our proposed directly trained token-level embeddings. --\textsuperscript{\textdagger}: Languages are not covered by the pretrained \fasttext{} word embeddings used by \wechsel{}.}
    \label{tab:init-mlm-comparison}
    \vspace{-2mm}
\end{table*}

We present downstream task results for NLI and QA in \autoref{tab:xnli-qa} and for NER in \autoref{tab:ner}. 
In the following, we discuss various aspects of these results.
In \autoref{fig:mlm-loss-comparison}, we show loss curves on a held out set when adapting XLM-R
with custom tokenizers.
In \autoref{tab:init-mlm-comparison}, we report the masked language modeling (MLM) loss of various methods right after initialization (no further training performed).

\paragraph{Effectiveness of \focus{}.} 
\autoref{tab:init-mlm-comparison} shows the effectiveness of \focus{} initialization for vocabulary replacement. Directly after initialization without further training, \focus{} significantly outperforms all other initialization methods. 
In \autoref{fig:mlm-loss-comparison}, we show loss curves over the course of language-adaptive pretraining (\lapt{}). For random initialization, we only plot the second stage after the embeddings have already been trained for an additional 20\% of total training steps. \focus{} yields a lower loss than random initialization even at the end of training, despite random initialization having been trained for more steps in total.
\wechsel{} starts off worse than \focus{} but catches up over the course of training.
Naturally, the effect of initialization is less pronounced the longer we train the models. We have deliberately constructed a difficult evaluation with our long training regime of 12.8 billion tokens.
In settings where less compute is available, \focus{} may be even more beneficial.

The improved effectiveness on the pretraining objective also translates to gains in downstream tasks, as reported in \autoref{tab:xnli-qa} and \autoref{tab:ner}.
\focus{} initialization outperforms random initialization across all downstream tasks and languages (except for Arabic TyDiQA). 
\wechsel{} also
improves over random initialization, but \focus{} obtains superior results.
\focus{} can also be applied for vocabulary extension instead of vocabulary replacement. Here, we see less of an improvement over the random initialization baseline.
This could be due to the smaller impact of \focus{}, since only a relatively small percentage of the large extended vocabulary is affected.

\paragraph{Vocabulary Extension or Replacement?} 
We find that vocabulary extension generally performs worse on downstream tasks than keeping the original vocabulary.
This finding is in line with results reported by \citet{ebrahimi-kann-2021-adapt} on a set of 30 typologically diverse languages. 
Prior studies proposing vocabulary extension \cite{language-adaptive-pretraing, wang-etal-2020-extending, chau-smith-2021-specializing} used mBERT and were motivated by the possibility of out-of-vocabulary (OOV) tokens. 
For \mbox{XLM-R} using SentencePiece with 100\% character set coverage or byte-level tokenizers, OOV tokens can always be represented at the character or byte level. 
Therefore, the benefits of vocabulary extension might be less pronounced in these cases because the OOV problem is less relevant to begin with. 

On average, when combined with \focus{} initialization, vocabulary replacement outperforms both vocabulary extension and keeping the original vocabulary.
Nevertheless, keeping the original vocabulary intact proves to be a strong baseline and for the high-resource language German even outperforms vocabulary replacement with \focus{}. 
However, vocabulary replacement paired with \focus{} performs better on 
medium- and low-resource languages, results in smaller models, and is thus faster to train. 

\paragraph{Low-Resource Languages.} Focusing on lesser-resourced languages, \focus{} outperforms random initialization and \lapt{} with the original vocabulary on NER for Hausa and isiXhosa.
Furthermore, we report the Masked Language Modeling loss directly after initialization on a number of very low-resource languages in \autoref{tab:init-mlm-comparison}. We see that across all languages, including the very low-resource ones, \focus{} achieves the best results. \focus{} also provides a good initialization even when the target language was not part of the source model's pretraining. 

In low-resource settings, a key advantage of \focus{} is that we only need unlabeled target language data to train our auxiliary embeddings -- a resource already needed for \lapt{} in any case. Unlike \wechsel{}, no bilingual dictionary is required, the quality and coverage of which might also be insufficient in low-resource settings. Some low-resource languages, such as Hausa and isiXhosa, are also not covered by \wechsel{}'s source of pretrained word embeddings.\footnote{\resizebox{0.93\linewidth}{!}{\url{https://fasttext.cc/docs/en/crawl-vectors.html}}}

\paragraph{Effect of Vocabulary Overlap.} Naturally, the quality and quantity of overlapping tokens influences the success of \focus{}. To investigate this, we conducted empirical analyses in two settings: using the full overlap and using only overlapping tokens that are symbols, numbers, or punctuation (Symbolic Overlap).
Full overlap can take advantage of the source model's multilingual pretraining if the target language or a closely related language were part of the pretraining corpus.
In any case, however, symbolic tokens such as whitespace, numbers, and punctuation should generally be available, allowing us to transfer a model to any language. 
In \autoref{tab:init-mlm-comparison}, we show that even when using only symbolic overlapping tokens, \focus{} outperforms \wechsel{} on medium to low-resource languages (e.g., Scottish Gaelic, Luxembourgish, Kiswahili, and others). For Arabic and German, \focus{} with only symbolic overlapping tokens performs slightly worse than \wechsel{}.
In practice however, we will generally have numerous further overlapping tokens such as named entities and code-switched tokens. This is demonstrated by our results for Luxembourgish, Cebuano, Samoan, and Hmong -- all languages that \mbox{XLM-R} and XLM-R's tokenizer were not pretrained on. Here, using the full overlap outperforms using only symbols, suggesting more beneficial overlapping tokens beyond the ones included in our symbolic overlap. 
Overall, these results show that \focus{} can provide a good initialization even when the target language was not part of the source model's pretraining.

\paragraph{Auxiliary Embeddings.} \wechsel{} proposes a method to use pretrained {word}-level \fasttext{} embeddings to obtain token-level embeddings. We propose to directly train token-level \fasttext{} embeddings. In \autoref{tab:init-mlm-comparison}, we additionally show \focus{}'s initialization performance when using the \wechsel{}-style method to obtain token-level \fasttext{} embeddings (Word-\fasttext{}). We see that using our directly trained token-level \fasttext{} embeddings results in a better initialization for low- and high-resource languages.

\paragraph{\wechselen{}.} On average, \wechsel{} actually fares slightly better than \wechselen{}, although \wechselen{} also improves over random initialization. 
For \wechselen{}, we followed \citet{wang-etal-2019-improving} in selecting English tokens in \mbox{XLM-R's} original vocabulary by taking the overlap with a language-specific English tokenizer's vocabulary.
Due to the substantial presence of English in \mbox{XLM-R's} original vocabulary, this may have been too restrictive, excluding too many potentially useful tokens.

\section{Related Work}
We now discuss further related work apart from the studies introduced in \autoref{sec:intro}. 
\paragraph{Language Adaptive Pretraining (\lapt{}).}
\citet{alabi-etal-2022-adapting} adapted XLM-R to up to 20 African languages at the same time instead of specializing on a single language.
\citet{ebrahimi-kann-2021-adapt} and \citet{wang-etal-2022-expanding} used resources with much higher language coverage than web-scraped monolingual texts (the Bible and lexicons, respectively) to adapt pretrained multilingual models to unseen languages. \citet{muller-etal-2021-unseen} transliterated unseen languages into Latin script to improve the results when using an existing pretrained vocabulary.

\paragraph{Adapters.}
In contrast to approaches changing all pretrained model weights, \citet{pfeiffer2020-madx} introduce additional adapter modules and only these new weights are changed. This is more parameter-efficient than full model adaptation, but gradients still need to be backpropagated throughout the model until the first adapter~\citep{adapterdrop}. Also, adapters introduce additional computational cost at inference time.

\paragraph{Bilingual Embedding Alignment.}
\citet{vernikos-popescu-belis-2021-subword-mapping} propose SMALA to calculate a mapping between embedding spaces for two languages to find semantically similar tokens across languages. They also experiment with initializing the remaining tokens based on this cross-lingual mapping.
\wechsel{} \cite{minixhofer-etal-2022-wechsel}
aligns word embeddings from two different languages. 
Such alignments operate under the assumption of near-isomorphism between embedding spaces of different languages~\cite{vulic-etal-2020-good}, i.e., that they share a similar geometric structure. Recent studies have challenged this assumption, especially for language pairs with typological~\cite{sogaard-etal-2018-limitations, patra-etal-2019-bilingual, ormazabal-etal-2019-analyzing} and resource~\cite{vulic-etal-2020-good,FuXianGengGeWangDongWangDeMelo2020ABSent} differences. This is especially detrimental in the case of language model transfer, as we usually transfer from a high-resource language such as English to less-resourced languages with potentially different typology. \focus{} does not require the alignment of embedding spaces.

For multilingual source models, \wechsel{} also disregards a valuable resource at our disposal: target language tokens that already have pretrained embeddings in the multilingual source model. For these tokens, we can copy their pretrained embeddings as a gold standard. Obtaining a different initialization is likely to lead to a worse result. \focus{} is well-positioned to take advantage of these pretrained embeddings of target language tokens.

Additionally, \wechsel{} requires a bilingual dictionary as an additional resource to seed the embedding space alignment. For low-resource languages, such a bilingual dictionary might be of lower quality or not available. 
\focus{} does not require bilingual dictionaries as an additional resource.

\paragraph{Other Embedding Initialization Methods.} In concurrent work, \citet{ostendorff2023efficient} propose a similar method to \focus{} that initializes an embedding matrix for a new vocabulary based on combinations of overlapping tokens with a pretrained embedding matrix, but use the embedding layer of a smaller pretrained Transformer model instead of static \fasttext{} embeddings as an auxiliary embedding space. 
However, their study only provides results on the high-resource language German as a target language and they do not consider BERT-style source models. 
If no smaller pretrained Transformer model with the desired tokenizer is available, training one from scratch comes with a much higher computational cost than training the \fasttext{} embeddings for \focus{}.
\citet{zeng2022greenplm} create a new vocabulary and embedding matrix for the target language by translating tokens in the source vocabulary with bilingual \mbox{dictionaries}.
\section{Conclusion}
\label{sec:conclusion}
We propose \focus{}, a novel embedding initialization method for the monolingual specialization of language models with a language-specific tokenizer. \focus{} uses the vocabulary overlap between source and target languages to effectively transfer the pretrained embeddings to the new target tokenizer's embedding matrix. 
In a series of experiments across a diverse set of languages and several different tasks,
we show that \focus{} outperforms other available embedding initialization methods without requiring additional resources like bilingual dictionaries. \focus{} can provide a good initialization even if only a minimal vocabulary overlap is available and when the target language has not been part of the source model's pretraining.
We release our \mbox{code and model checkpoints on GitHub}.\footnote{\url{https://github.com/konstantinjdobler/focus}}

\section*{Limitations}
We evaluate \focus{} only for BERT-like Transformer models. 
In principle, the method is applicable to any model that uses a vocabulary, tokenizer, and embedding matrix. 
In future work, we hope to investigate the use of \focus{} on GPT decoder models, as explored by \citet{ostendorff2023efficient}.

We conduct downstream task evaluations for NLI, QA, and NER on German, Arabic, and Swahili. For the low-resource languages isiXhosa and Hausa, we conduct downstream task experiments for NER.
This provides a good mix of different levels of available resources, scripts, and typology.  
However, further evaluations on languages covering more scripts and languages that were not part of the source models' pretraining are needed to substantiate the effectiveness of \focus{} in these settings.
All our chosen languages have monolingual texts available for further pretraining.
As \citet{wang-etal-2022-expanding} note, this is not the case for many other low-resource languages. 
Since further pretraining on target language data is a key component of our model adaptation strategy, the applicability of \focus{} is also limited in this regard, although such data can in some cases also be synthesized.

\section*{Ethics Statement}
In this work, we conduct the main part of our downstream task experiments on German, Arabic, and Swahili. These choices stem from our desire to provide practically useful ideas that reflect the current availability of models and to conduct experiments on downstream tasks such as question answering, NLI, and named entity recognition, for which we need relevant ground truth data. 

Finally, researchers and practitioners need to be cognizant of the fact that adopting existing monolingual or even multilingual models as a starting point instead of training new models from scratch can lead to remnant biases towards the original pretraining data. Hence, there is a risk that a model adopts certain forms of behavior that reflect other languages and cultures than that of the language community one is targeting. Also, web-scale datasets used for pretraining such as CC100 might contain personal and sensitive information. Such behavior needs to be assessed very carefully before any real-world deployment of the models.

\section*{Acknowledgements}
The authors acknowledge the financial support by the German Federal Ministry for Education and Research (BMBF) through the project <<KI-Servicezentrum Berlin Brandenburg>> (01IS22092). We also thank the reviewers for their helpful comments.

\bibliography{anthology,custom}
\bibliographystyle{acl_natbib}

\cleardoublepage
\appendix

\section{Hyperparameters and Experiment Details}
\label{app:hyperparams}
We conducted all experiments on a heterogeneous compute cluster with Nvidia V100 32GB, A100 40GB, A100 80GB, and A6000 48GB GPUs. Depending on availability we used one, two, or four GPUs for our experiments and adjusted the batch size per device so that we retain the same effective batch size. 
Depending on total model size, we also used gradient accumulation with smaller batch sizes to fit the model into GPU memory. We used PyTorch~\cite{pytorch} and \texttt{pytorch-lightning}~\cite{Falcon_PyTorch_Lightning_2019} as well as the HuggingFace \texttt{transformers}~\cite{Wolf_Transformers_State-of-the-Art_Natural_2020}, \texttt{tokenizers}~\cite{tokenizers} and \texttt{datasets}~\cite{Lhoest_Datasets_A_Community_2021} libraries. We used the \texttt{fp16} mixed precision training implemented by \texttt{pytorch-lightning}.

\paragraph{Further pretraining.}
\phantomsection
\label{app:hyperparams:par:further-pretraining}
We used the same hyperparameters for all target languages, as detailed in \autoref{tab:params-pretraining}. We trained for a total of 50 million samples with batches of 128 sequences of 256 tokens. This results in a total of 390,625 optimizer steps (weight updates). We used AdamW optimization~\cite{loshchilov2017decoupled} as implemented in \texttt{torch.optim}\footnote{Since we do not use weight decay, this is equivalent to using Adam~\cite{adam}.} and a linear learning rate warmup for 5 million samples (39,062 optimizer steps) followed by a constant learning rate at $5 \times 10^{-5}$. We used a constant schedule to allow for more flexible experimentation regarding the total number of training steps and to ensure that the impression of converged loss curves is not a false positive induced by a decaying learning rate. We also conducted preliminary experiments using a cosine learning rate schedule and did not observe a significant difference.
We used the CC100 dataset in line with its intended use for the pretraining of language models. For German, our training does not even complete a full epoch. 

\paragraph{Datasets for low-resource languages.} Comparing OSCARv23.01 and mC4 as a data source for  low-resource languages, we observe that for Corsican\footnote{Results not included in the paper as the language is not provided by OSCARv23.01.}, Cebuano and Luxembourgish, the quality in mC4 is quite poor. Training a tokenizer on these datasets results in a tokenizer that, on average, encodes \textit{fewer} characters per token than when using the original XLM-R tokenizer. Motivated by this, we turned to OSCARv23.01 (for Cebunao and Luxembourgish, since it does not contain Corsican). We filter each dataset based on the quality warnings \texttt{header}, \texttt{footer}, and \texttt{noisy} provided by OSCARv23.01. Training a tokenizer on this data yielded the expected results.
For Hmong and Samoan, we did not observe such degraded tokenization training on mC4. Nevertheless, these corpora (and all web-crawled corpora of low-resource languages) can also be expected to be noisy.

\begin{table}

    \begin{tabular}{lcccccc}
    \toprule
    Hyper-parameter & Value\\
    \midrule
    peak learning rate & $5 \times 10^{-5}$\\
    learning rate schedule & constant\\
    learning rate warmup & 5 million samples\\
    batch size & 128\\
    sequence length & 256\\
    gradient clipping & 1.0\\
    Adam $\epsilon$ & $1 \times 10^{-8}$\\
    Adam $\beta_1$ & 0.9 \\
    Adam $\beta_2$ & 0.999 \\
    training samples & 50 million\\
    resulting train steps & 390625\\
    \bottomrule
    \end{tabular}
    \centering
           \caption{Hyper-parameters for further pretraining on target language data.}
               \label{tab:params-pretraining}
\vspace{2mm}
\end{table}

\paragraph{Downstream tasks.}
\phantomsection
\label{app:hyperparams:par:downstream}
We detail our hyperparameters for all downstream tasks in \autoref{tab:params-downstream}. We largely followed default values of finetuning scripts provided by Huggingface\footnote{\href{https://github.com/huggingface/transformers/tree/main/examples/pytorch}{{\texttt{github.com/huggingface/transformers/tree/main/\\examples/pytorch}}}}, but adjusted the training epochs depending on dataset size, added a linear learning rate warmup for 10\% of total training steps, and adjusted the batch size based on used GPU memory per task. Additionally, we used a $2 \times 10^{-5}$ peak learning rate for all non-QA tasks. We repeated each experiment five times with the random seeds \texttt{\{1,2,3,4,5\}} and report the mean and standard deviation across runs. For XNLI, we report accuracy, for TyDiQA, GermanQuAD, WikiANN, MasakhaNERv2, and GermEval14, we report the F$_1$-Score.

\paragraph{Tokenizer training.} 
\phantomsection
\label{app:hyperparams:par:tokenizer}
Our language-specific tokenizers were trained in the same way as XLM-R for comparability, specifically SentencePiece tokenization~\cite{sentencepiece} with the Unigram algorithm~\cite{unigram2018kudo}. We used HuggingFace \texttt{tokenizers} and a vocabulary size of 50k tokens for all languages.
The resulting vocabularies contain a large amount (roughly 10k tokens) of emojis and Chinese, Japanese, and Korean single characters, which is an artifact of SentencePiece's \texttt{character\_coverage} parameter (which defaults to 100\%). Characters are included in the vocabulary even if they appear only once in the large amount of noisy web-scraped training documents.
This effectively means that our language-specific vocabularies are roughly 10k tokens smaller in practice, as such single characters rarely occur in the training data. 
In practice, one may wish to tune the \texttt{character\_coverage} carefully based on the requirements of the target language if a smaller model is desired.

\begin{table*}
    \begin{tabular}{lcccccc}
    \toprule
    Hyper-parameter & XNLI & QA & WikiANN & GermEval14 & MasakhaNERv2 \\
    \midrule
    epochs & 2 & 5* & 5* & 25 & 25\\
    peak learning rate & $2 \times 10^{-5}$ & $ 5\times 10^{-5}$ & $ 5\times 10^{-5}$ & $ 5\times 10^{-5}$ & $ 2\times 10^{-5}$ \\
    lr schedule & linear & linear & linear & linear & linear \\
    lr warmup ratio & 10\% & 10\% & 10\%& 10\%& 10\%\\
    batch size & 128 & 64 & 128 & 128 & 128\\
    sequence length & 256 & 384 & 256 & 256 & 256\\
    gradient clipping & 1.0 & 1.0 & 1.0& 1.0& 1.0\\
    Adam $\epsilon$ & $1 \times 10^{-8}$ & $1 \times 10^{-8}$ & $1 \times 10^{-8}$& $1 \times 10^{-8}$& $1 \times 10^{-8}$\\
    Adam $\beta_1$ & 0.9 & 0.9& 0.9& 0.9& 0.9 \\
    Adam $\beta_2$ & 0.999 & 0.999 & 0.999 & 0.999 & 0.999  \\
    \bottomrule
    \end{tabular}
    \centering
           \caption{Hyper-parameters for our downstream tasks. *: After observing high variance due to smaller training set sizes, we adjusted the number of epochs for Swahili to 25 for QA (on TyDiQA GoldP) and NER (on WikiANN). We evaluate every 5\% of total training steps and report the best checkpoint's results on the test set.}
    \label{tab:params-downstream}
\end{table*}

\section{Further details for \focus{}}
\label{app:further-focus-details}
\paragraph{FastText training.}
To obtain static token embeddings for \focus{}, we train \fasttext{} embeddings on tokenized target language training data. We mostly used default hyper-parameters but increased the dimensionality to 300, as is commonly done in the literature~\citep{Mikolov2013word2vecO,bojanowski-etal-2017-enriching}. We ran the training for three epochs. On German, due to its corpus size, we ran only a single epoch. 

Additionally, we set a \texttt{minCount} of 10 for tokens during \fasttext{} training to filter out very rare tokens. These rare tokens are initialized from a normal distribution with mean and standard deviation per dimension calculated from the source embedding matrix, as done by \wechsel{} for tokens that have no subwords in the pretrained word embedding.
Setting \texttt{minCount} also helps with filtering the noisy single characters that are part of our tokenizers due to SentencePiece's \texttt{character\_coverage} parameter. 

\paragraph{Vocabulary Overlap.}
\phantomsection
\label{app:fasttext:par:vocab}
\begin{table*}[t]
    \centering
    \begin{tabular}{@{}lcc@{}}
    \toprule
    Category & Share & Examples\\
    \midrule
    Symbols \& Numbers & \hphantom{0}9\% & 3.5, 1919, 3500, ..., ;-), !!!, 1\%, [6]\\
    Names \& Entities & 10\% & BlackBerry, Oscar, Messi, JavaScript\\
    German (sub-)words  & 46\% & Bewerber, Wahl, günstig, fallen\\
    English \& Code-switched & 18\% &  Smoothie, FAQ, Backup, Settings \\
    Not assignable & 17\% & ik, Kri,  kub, rez, zy, BF, oka \\
    \bottomrule
    \end{tabular}
    \caption{Investigation of overlapping tokens on German. The evaluation was conducted manually by one of the authors on a random sample of 500 overlapping tokens. In the examples, leading spaces are omitted and we further exclude the ``noisy'' single-character tokens mentioned in Appendix~\ref{app:hyperparams:par:tokenizer}.}

    \label{tab:overlapping-token-sherlock}
\end{table*}
\begin{table}
    \centering
    \begin{tabular}{lrr}
    \toprule
    & \multicolumn{2}{c}{Tokens in Overlap}\\
    \cmidrule(lr){2-3}
    Language &  \texttt{minCount} = 10 & Full \\
    \midrule
    German & 14,485 & 18,986\\
    Arabic &  10,658 & 13,996\\
    Swahili & 10,443 & 12,353 \\
    Hausa & 11,481 & 14,806 \\
    isiXhosa & 6,222 & 8,333\\
    \bottomrule
    \end{tabular}
    \caption{Number of tokens in the overlap between language-specific and XLM-R's original vocabulary. For learning \fasttext{} embeddings with \focus{}, we set \texttt{minCount} = 10, which filters out very rare and noisy tokens.}
    \label{tab:overlapping-size}
\end{table}

\focus{} relies on overlapping tokens between the new and pretrained vocabularies.
Ideally, an overlapping token would have the same semantics in the target language vocabulary and in the pretrained vocabulary. 
If the target language was already part of the pretraining, this is most obviously true for (sub-)words that only occur in the target language. Differences in script or peculiarities of the target language (such as German umlauts and other language-specific accented characters)
help facilitate such occurrences. 
In many languages, especially online, there is widespread code-switching with English, leading to English words being interspersed within native sentences, which also contributes to shared semantics. 
A considerable share of tokens is also made up of names, named entities, symbols, numbers, and punctuation. 
While these are not exclusive to any particular language, they are likely to possess the same semantics {across} languages, making them good overlapping tokens. We report the number of overlapping tokens for languages used during training in \autoref{tab:overlapping-size}.

Additionally, we manually classified a random sample of 500 overlapping tokens for German and report the results in \autoref{tab:overlapping-token-sherlock}. The overlap is calculated between XLM-R's original tokenizer and our newly trained, language-specific German one. For this analysis, we excluded the noisy single-character tokens mentioned in Appendix~\ref{app:hyperparams:par:tokenizer}.
We conclude that a considerable share of the overlapping tokens for German does indeed possess similar semantics in the pretrained and new vocabularies. 
For less-resourced languages than German that were still part of the multilingual models' pretraining, we can expect fewer overlapping tokens that are directly part of the target language. High-resource languages have a larger share of language-specific tokens in the vocabulary of XLM-R.
However, for languages with an uncommon or unique script, tokens are more likely to be exclusive to the target language.
During the pretraining of XLM-R, low-resource languages are also oversampled~\cite{xlmr}. 
Therefore, tokens that are shared between low and high-resource languages are more likely to also have the low-resource language semantics encoded in their embeddings than would otherwise be the case.

\paragraph{Overlaps between different tokenizers.}
In general, we only consider tokens as overlapping if they are an exact match (including case and the ``beginning of word'' (\textsc{Bow}) signifier. However, for tokens that only consist of numbers, punctuation, or whitespace, we implement fuzzy matching where we disregard the case and the \textsc{Bow} signifier.

A peculiarity of calculating token overlaps between different \textit{kinds} of tokenizers is the representation of tokens that are \textsc{Bow} tokens and non-ASCII characters. For example, the HuggingFace implementation of Byte-Level BPE uses \texttt{Ġ} as a prefix for \textsc{Bow} tokens, whereas XLM-R's tokenizer \mbox{uses \texttt{\_}}. To complicate things, BERT's tokenizer WordPiece~\cite{devlin-etal-2019-bert} prefixes tokens that are \textit{not} \textsc{Bow} with \texttt{\#\#}. Also, Byte-level BPE represents non-ASCII characters in tokens differently than XLM-R's tokenizer. In our experiments in this paper, we only use the XLM-R tokenizer, which also matches the source model's tokenizer, and therefore avoid these problems. However, a correct canonicalization of tokens to a common form is crucial to enable \focus{} 
when the tokenizers of source and target model do not match. We implement such a canonicalization method for common tokenizers and release it as part of our ready-to-use implementation of \focus{}.\footnote{\url{https://github.com/konstantinjdobler/focus}}

\end{document}